\documentclass[acmtog]{acmart}
\acmSubmissionID{594}

\usepackage{booktabs} % For formal tables

\usepackage[ruled]{algorithm2e} % For algorithms

\usepackage{caption}
\usepackage{subcaption}
\usepackage{multirow}
\usepackage{bm}

\SetAlFnt{\small}
\SetAlCapFnt{\small}
\SetAlCapNameFnt{\small}
\SetAlCapHSkip{0pt}

\setcopyright{acmcopyright}
\acmJournal{TOG}
\acmYear{2022}\acmVolume{41}\acmNumber{4}\acmArticle{140}\acmMonth{7}
\acmDOI{10.1145/3528223.3530159}

\begin{document}
% Title portion
\title{DCT-Net: Domain-Calibrated Translation for Portrait Stylization}

% DO NOT ENTER AUTHOR INFORMATION FOR ANONYMOUS TECHNICAL PAPER SUBMISSIONS TO SIGGRAPH 2019!
\author{Yifang Men}
\orcid{0000-0003-2495-2869}
\authornote{Corresponding authors.}
\affiliation{%
  \institution{DAMO Academy, Alibaba Group}
  %\city{Beijing}
  \country{China}}
\email{myf272609@alibaba-inc.com}

\author{Yuan Yao}
\affiliation{%
  \institution{DAMO Academy, Alibaba Group}
   % \city{Beijing}
  \country{China}}
\email{yaoy92@gmail.com}

\author{Miaomiao Cui}
\affiliation{%
  \institution{DAMO Academy, Alibaba Group}
   % \city{Beijing}
  \country{China}}
\email{miaomiao.cmm@alibaba-inc.com}

\author{Zhouhui Lian}
\authornotemark[1]
\affiliation{%
 \institution{Wangxuan Institute of Computer Technology, Peking University}
  \city{Beijing}
 \country{China}}
\email{lianzhouhui@pku.edu.cn}

\author{Xuansong Xie}
\affiliation{%
  \institution{DAMO Academy, Alibaba Group}
   % \city{Beijing}
  \country{China}}
\email{xingtong.xxs@alibaba-inc.com}

\renewcommand\shortauthors{Men, Y. et al}

\begin{abstract}
This paper introduces DCT-Net, a novel image translation architecture for few-shot portrait stylization. Given limited style exemplars ($\sim$100), the new architecture can produce high-quality style transfer results with advanced ability to synthesize high-fidelity contents and strong generality to handle complicated scenes (e.g., occlusions and accessories). Moreover, it enables full-body image translation via one elegant evaluation network trained by partial observations (i.e., stylized heads). Few-shot learning based style transfer is challenging since the learned model can easily become overfitted in the target domain, due to the biased distribution formed by only a few training examples. This paper aims to handle the challenge by adopting the key idea of ``calibration first, translation later'' and exploring the augmented global structure with locally-focused translation. Specifically, the proposed DCT-Net consists of three modules: a content adapter borrowing the powerful prior from source photos to calibrate the content distribution of target samples; a geometry expansion module using affine transformations to release spatially semantic constraints; and a texture translation module leveraging samples produced by the calibrated distribution to learn a fine-grained conversion. Experimental results demonstrate the proposed method’s superiority over the state of the art in head stylization and its effectiveness on full image translation with adaptive deformations. Our code is publicly available at https://github.com/menyifang/DCT-Net. 
\end{abstract}






\begin{teaserfigure}
    \centering
   \setlength{\abovecaptionskip}{-0.1cm}
  \includegraphics[width=0.98\textwidth]{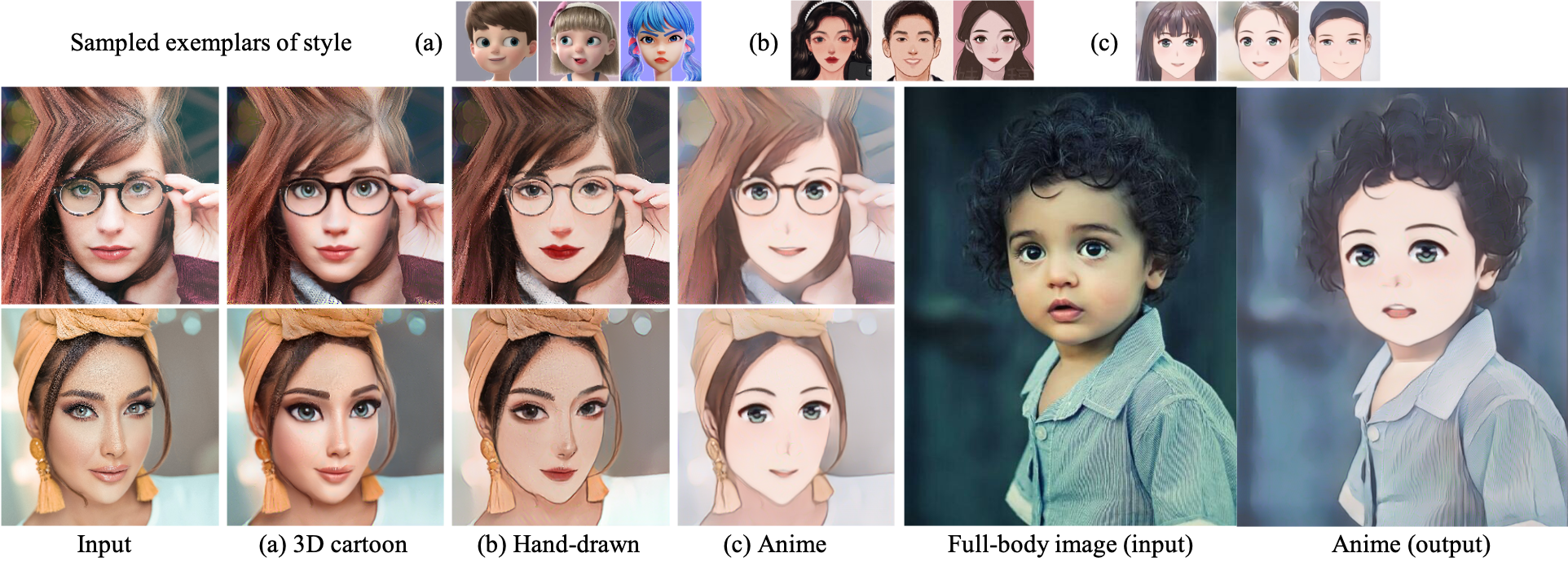}
  \captionof{figure}{Given limited style exemplars, our method can synthesize artistic portraits in corresponding styles, excelling in content (e.g., identity and accessories) preservation, and handling complicated faces with heavy occlusions, makeup, or rare poses. Our method also enables full-body image translation by using only head observation for training samples. Source credits: head input and full-body input \copyright Pexels website.}
    \label{fig:teaser}
\end{teaserfigure}

\maketitle

\section{Introduction}

Portrait stylization, an essential part of digital art, aims to transform natural persons' appearances into more creative interpretations in desired visual styles while maintaining personal identity. It changes source portraits with beautified or exaggerated effects in a fantastic way and has enormous potential applications including art creation, animation making, and virtual avatar generation. However, creating artistic portraiture is skill-restrictive and requires substantial human labors for image creation and arrangement. 
%Stylized portrait, an essential part in digital art, aims at transforming the appearance of real persons into more creative interpretations in desired visual styles while keeping the personal identity unchanged. It changes source portraits with beautified or exaggerated effects in a fantastic way and has huge potential applications for art creation, animation making and virtual avatar generation. However, creating an art portraiture is skill-restrictive and requires substantial human labor for image creation and arrangement. 

With the rapid development of Generative Adversarial Networks (GANs)~\cite{goodfellow2014generative, mirza2014conditional}, image-to-image translation methods~\cite{isola2017image, wang2018high} have been introduced to automatically learn a function that maps images from one domain to the other. Due to the unavailability of paired data, existing methods~\cite{zhu2017unpaired, chen2018cartoongan, chen2019animegan, Kim2020U-GAT-IT} mainly utilize cycle consistency to learn a translation from source photos to cartoonized results. However, these methods still require a large amount of unpaired data and easily suffer from notable texture artifacts in complex scenes. Recently, stylizing faces by leveraging the pre-trained StyleGAN~\cite{karras2019style,karras2020analyzing} has gained intensive attention~\cite{pinkney2020resolution,richardson2021encoding, song2021agilegan}. Compared to previous conditional generative models, they make full use of the powerful generative capability of unconditional StyleGAN, thus producing high-quality portraits with limited style exemplars. 
Due to the nature of unconditional generation, they typically learn a cartoon generator to map random noises to cartoon images, and combine inversion algorithms using optimization~\cite{abdal2019image2stylegan,abdal2020image2stylegan++,creswell2018inverting} or learning based methods~\cite{perarnau2016invertible,bau2019inverting} to project real photos into latent codes in the StyleGAN space, thus achieving the photo-to-cartoon translation. Despite high-quality results produced, these methods often suffer from the content missing problem owing to the limited generalization ability of arbitrary out-of-domain faces. Moreover, all these existing methods are tailored for heads and can not handle full-body images.

\begin{figure}
\begin{center}
\setlength{\abovecaptionskip}{0cm}
\setlength{\belowcaptionskip}{-0.1cm}
\includegraphics[width=1.0\linewidth]{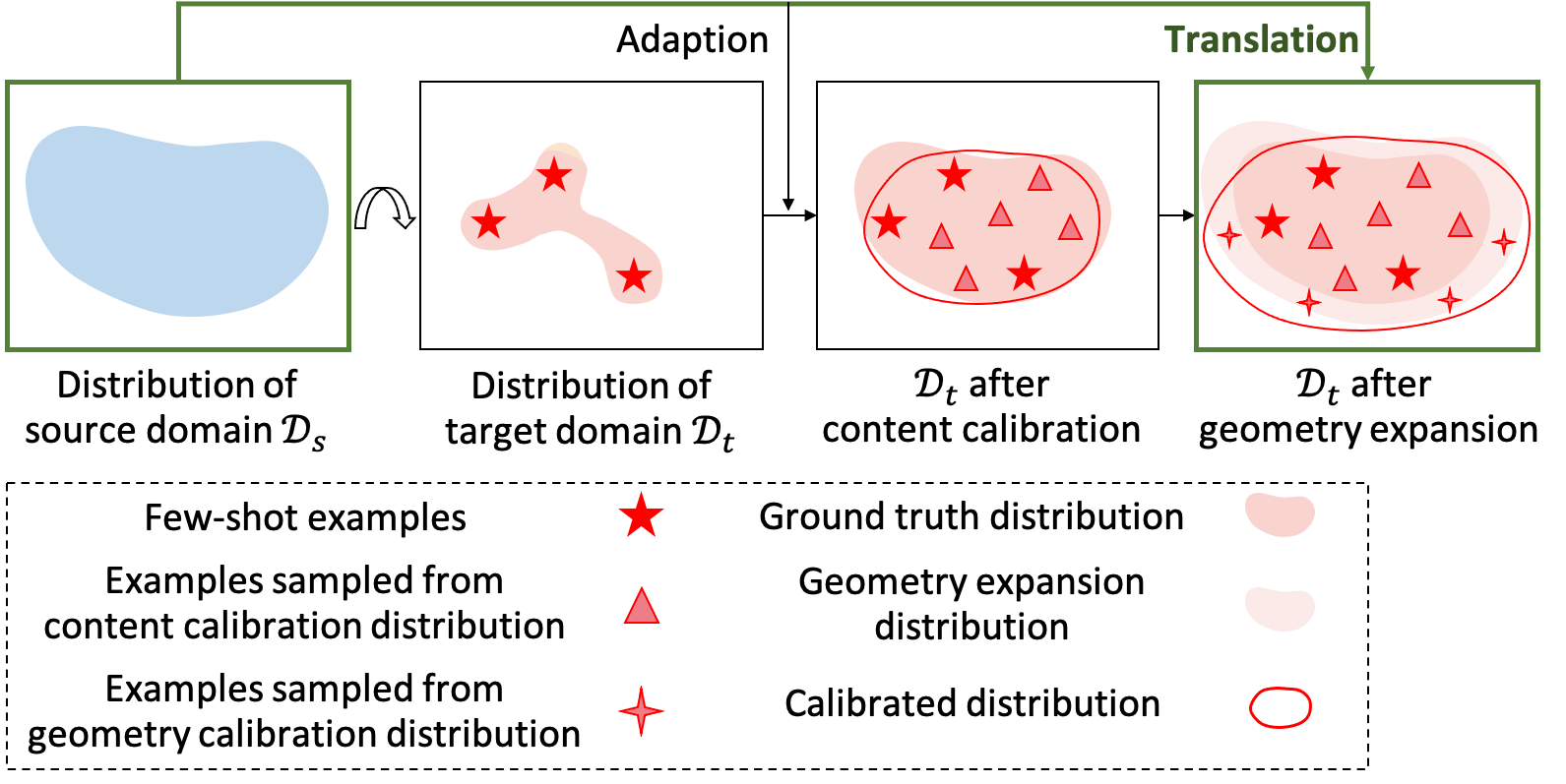}
\caption{An illustration of domain-calibrated translation. It is difficult to learn correspondences from the diverse source distribution to the biased target distribution formed by few-shot examples. We firstly calibrate the distribution of the target domain $\mathcal{D}_t$ in content features by adapting source samples, and then expand $\mathcal{D}_t$ in the geometry dimension. With examples sampled from the calibrated distribution, it is easier to learn a fine-grained texture translation with advanced ability, generality, and scalability. }
\label{fig:insight}
\end{center}
\end{figure}

The aim of this paper is to propose a new and effective method for portrait stylization, which can simultaneously achieve advanced ability to synthesize high-preserving contents, strong generality to handle complicated real-world scenes, and high scalability to transfer various styles. As depicted in Figure~\ref{fig:teaser}, given a small amount of style exemplars ($\sim$100), our method can translate arbitrary real faces to artistic portraits in corresponding styles (e.g., 3D-cartoon, anime, and hand-drawn), even full-body images can be properly processed with adaptive deformations (e.g., exaggerated facial features and faithful body textures). Due to the insufficient and partial observation (only head regions) of style exemplars as well as the diversity of real-world scenes, it is challenging to achieve the goal mentioned above. Rethinking the essence of the task, it actually tries to learn cross-domain correspondences from the diverse source distribution to the biased target distribution formed by only a few training examples, as illustrated in Figure~\ref{fig:insight}. The learned model can easily become overfitted in the target domain and thus generating unsatisfactory style transferring results.

% %%%%%%%%%%%% figure 2

%{\bf Contributions:}  
The \emph{key insights} of this paper are threefold. First, the ``calibration first, translation later'' strategy makes it easier to learn stable cross-domain translation and produce high-fidelity results. Second, the balanced source distribution can be used as a prior to calibrate the biased content distribution of the target domain. Third, releasing spatially semantic constraints via geometry expansion leads to more flexible and wider-range inference. 
To this end, we propose a simple yet effective solution for ``domain-calibrated translation'', which firstly calibrates the content features of the target distribution by adapting the learned source generator to the target domain (i.e., borrowing the powerful content prior from source). Then, the domain features are further enriched using a geometry expansion module. With these calibrated distributions, an adequate number of diverse examples depicting non-local correlation can be produced, and we train a U-net, a network with strong local behavior, to perform the cross-domain translation. This design
makes our method be capable of learning the augmented global structure with locally-focused translation and brings all-around improvements. Our trained model excels in not only preserving detailed contents (e.g., identity, accessories, and backgrounds), but also handling complex scenes (e.g., heavy occlusions and rare poses). It also greatly increases the translation's generalization capabilities, allowing out of domain translations, such as full-body image translation. This brand-new task requires adaptive deformations when only trained on raw head collections. To the best of our knowledge, this is the first approach to propose the structure of ``domain-calibrated translation'' and show its superiority in the above aspects.

%%%%%%%%%%%%%%%%%%%%%%%%%%%%%%%%%%%%%%%%%%%%% Related work
\section{Related Work}
\subsection{Neural Style Transfer.}
Style transfer is a kind of non-realistic rendering technique~\cite{kyprianidis2012state}. Inspired by the power of CNN, ~\cite{gatys2015texture, gatys2016image} opened up a new field named Neural Style Transfer (NST), which presents an optimization based method for transferring the style of a given artwork to an image. Several works that target portraits thereafter specifically achieved impressive results. ~\cite{selim2016painting} proposed a head portrait painting method by locally transferring the color distributions of the example painting to others. ~\cite{kaur2019photo} devised a method to transfer face texture from a style face image to a content face image in a photo-realistic manner. However, these methods are closely related to texture synthesis and fail to handle geometric transformations.

% %%%%%%%%%%%% figure 3
\begin{figure*}
\begin{center}
\setlength{\abovecaptionskip}{-0.1cm}
\includegraphics[width=0.83\linewidth]{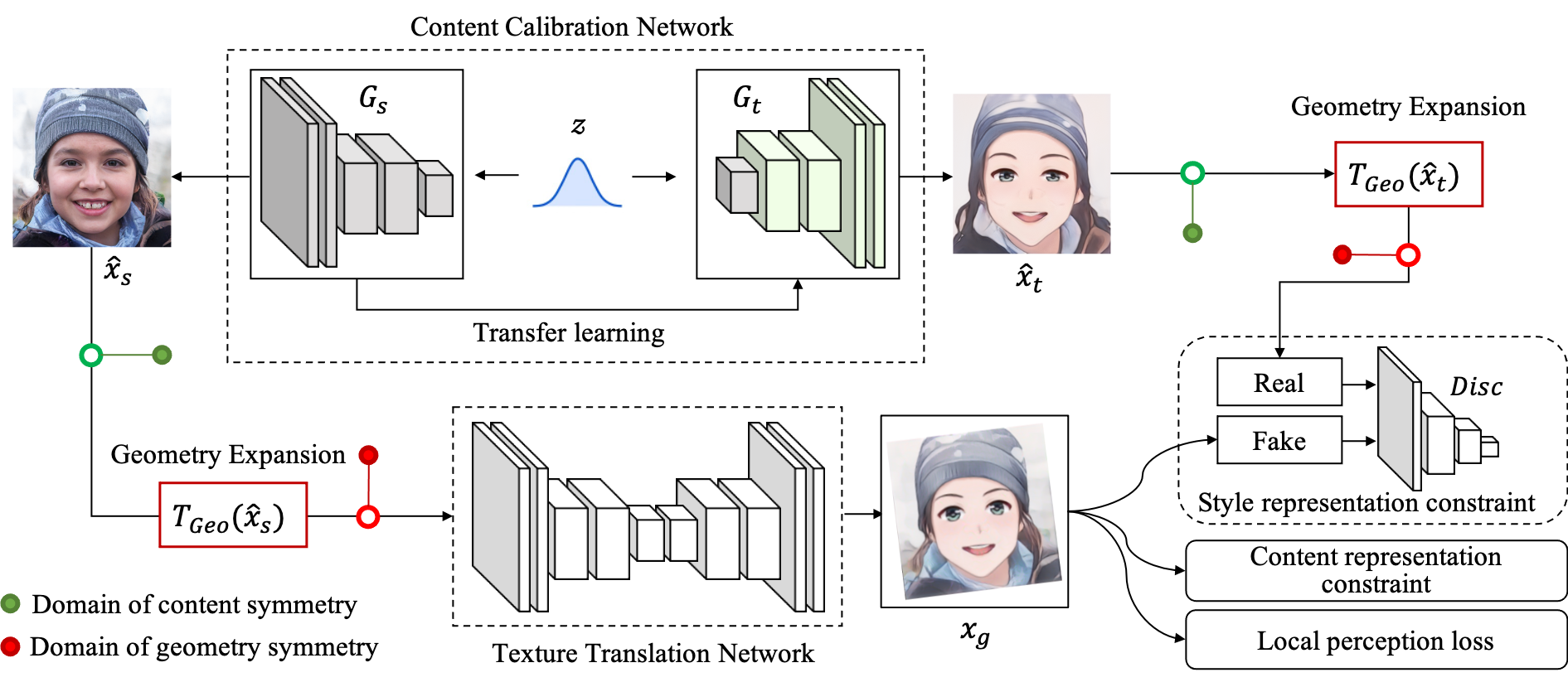}
\caption{ An overview of the proposed framework, which consists of the content calibration network (CCN), the geometry expansion module (GEM), and the texture translation network (TTN). CCN borrows the content prior from the real face generator $G_s$ and adapts it to the target domain, thus calibrating the content distribution of the target domain and obtaining content symmetric features. GEM expands the geometry distribution of two domains to release spatial constraints and enhance geometry symmetry. With calibrated domains, TTN is adopted to learn cross-domain translation with multi-representation and local perception constraints. CCN and TTN are trained independently. After training, only TTN is used for the final inference.}
\label{fig:pipeline}
\end{center}
\end{figure*}

\subsection{Image-to-Image Translation.} It aims to learn a mapping between images in two different domains. \cite{isola2017image} first proposed a supervised image translation model with conditional GANs~\cite{mirza2014conditional}, and was later extended to synthesize high-resolution images~\cite{wang2018high}. To alleviate the difficulty of acquiring paired data, ~\cite{zhu2017unpaired} proposed a cycle-consistency loss to use unpaired data for the translation task. A number of variants~\cite{liu2017unsupervised,huang2018multimodal,choi2020stargan} have been developed thereafter to adapt this framework to different scenarios. Despite the utilization of dedicated architectures designed in aforementioned methods, their abilities of generalizing to discrepant domains are restricted. There exist some works~\cite{cao2018carigans,shi2019warpgan,gong2020autotoon} that apply this framework to learn both texture and geometric styles for caricatures generation. Exaggerations learned in this task rely on local warping features, which is restricted for specific style transfer. Recently, ~\cite{Kim2020U-GAT-IT} incorporated a new attention module and a new learnable normalization function for unsupervised image translation tasks, which enables performing translation for requirement of both holistic and large shape changes. Nevertheless, it still requires extensive unpaired training data and easily generates unstable results.

\subsection{GAN Inversion.} In order to support real image editing with pretrained GANs, a specific task known as GAN Inversion, is used to learn the natural image manifold and inversely manipulate images into the latent space of a GAN model. Generally, there are three main techniques of GAN inversion~\cite{xia2021gan}, i.e., projecting an image into the corresponding latent space based on learning~\cite{zhu2016generative,perarnau2016invertible,bau2019inverting}, optimization~\cite{abdal2020image2stylegan++,ma2019invertibility}, and hybrid formulations~\cite{bau2019seeing,zhu2020domain}. With a novel style-based architecture, StyleGAN~\cite{karras2019style,karras2020analyzing} has been shown to contain a semantically rich latent space that can be used for inversion tasks. Recently, ~\cite{viazovetskyi2020stylegan2} distilled StyleGAN2 into the image-to-image network in a paired way for face editing. ~\cite{richardson2021encoding} proposed a generic Pixel2Style2Pixel (PSP) encoder to extract the learned styles from
the corresponding feature map, and can further be used to solve image-to-image translation tasks such as inpainting, super resolution, and portrait stylization. ~\cite{tov2021designing} designed a new encoder to facilitate higher editing quality on real images. 
~\cite{pinkney2020resolution} proposed a GAN interpolation framework for controllable cross-domain image synthesis, allowing to generate the ``Toonified'' version of the original image. More closely related to our approach is AgileGAN~\cite{song2021agilegan}, which introduces an inversion-consistent transfer learning framework for high-quality stylistic portraits, 
A later work ~\cite{ojha2021few} tried to generate stylized paintings using few-shot exemplars via cross-domain correspondence. However, all these works suffer from the content missing problem and can not tackle hard cases in real images (e.g., accessories and occlusions), due to the weakness of out-of-distribution generalization ability. In contrast, we present a novel domain-calibrated translation framework to well adapt the original training distribution.

%%%%%%%%%%%%%%%%%%%%%%%%%%%%%%%%%%%%%%%%%%%%% Method Description
\section{Method Description}
\subsection{Overview}
Given a small set of target stylistic exemplars, our goal is to learn a function $M_{s\to t}$ that maps images from the source domain $X_s$ to the target domain $X_t$. The output image $x_g$ should be rendered in the similar texture style of the target exemplar $x_t$, while preserving the content details (e.g., structure and identity) of the source image $x_s$. 

An overview of the proposed framework is shown in Figure~\ref{fig:pipeline}. We build a sequential pipeline with the following three modules: the content calibration network (CCN), the geometry expansion module (GEM), and the texture translation network (TTN). The first module is responsible for calibrating the target distribution in the content dimension by adapting the target style from a pre-trained source generator $G_s$ with transfer learning. The second module further expands the geometry dimension of both source and target distributions, and provides geometry-symmetry features with different scales and rotations for the later translation. With data sampled from the calibrated distribution, our texture translation network is employed to learn cross-domain correspondences with multi-representation constraints and the local perception loss. CCN and TTN are trained independently, and only TTN is used for the final inference. In the following, we will give a detailed description for each module of our framework.

\subsection{Content calibration network}
\begin{figure}
\begin{center}
\setlength{\abovecaptionskip}{0.1cm}
\includegraphics[width=0.92\linewidth]{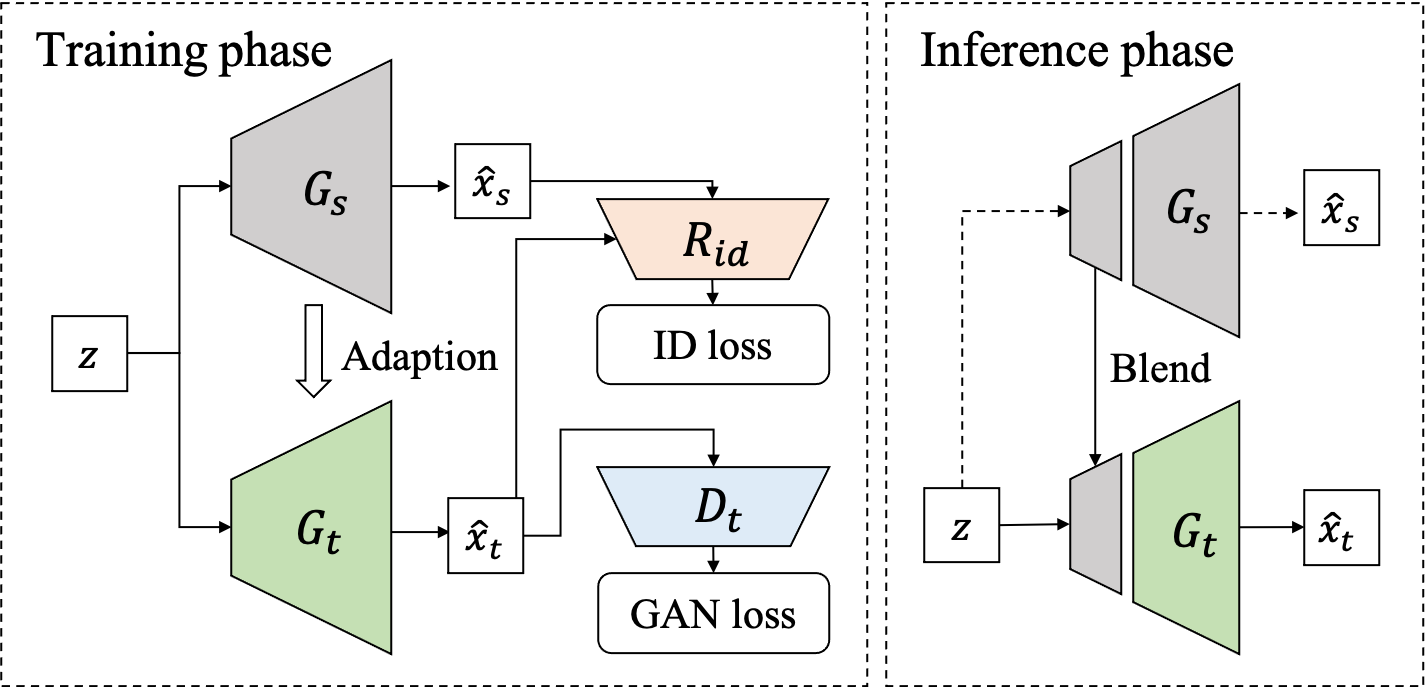}
\caption{The flowchart of the content calibration network.}
\label{fig:ccn}
\end{center}
\end{figure}

In this module, we calibrate the biased distribution of a few target samples by transferring network parameters learned from sufficient examples.
Different from pervious works~\cite{pinkney2020resolution,richardson2021encoding, song2021agilegan} combining StyleGAN2~\cite{karras2020analyzing} with inversion methods for image translation, we leverage the powerful prior from pre-trained StyleGAN2 to reconstruct the target domain with enhanced content symmetry. Starting from a StyleGAN2-based model $G_s$ trained on real faces (e.g., the FFHQ dataset), $G_t$, a copy of $G_s$, is used as initialization weights and we adapt $G_t$ to generate images in the target domain $X_t$. During the training phase of CCN, we fine-tune $G_t$ with a discriminator $D_t$ to ensure $\hat{x}_t \in X_t$ and an existing face recognition model $R_{id}$~\cite{deng2019arcface} to preserve the person identity between $\hat{x}_t$ and $\hat{x}_s$. During the inference phase of CCN, we blend the first $k$ layers of $G_s$ with the corresponding layers of $G_t$, which has been proven to be effective to preserve more contents of the original source domain~\cite{pinkney2020resolution}. In this way, we can produce relatively content-symmetric images in source and target domains, such as $\hat{x}_s$ and $\hat{x}_t$. The flowchart is displayed in Figure~\ref{fig:ccn}.

% %%%%%%%%%%%% figure 4

It is worthy of noting that we directly sample from the $z$ space and reconstruct the source and target domains $( \hat{X}_s, \hat{X}_t )$ in a content-symmetric way (i.e., the same $z$ for two decoding pathways). No real faces are used which need inversion embedding and lead to accumulated errors. Due to sufficient data of real-world photos, the distribution $\mathcal{D}(\hat{X}_s)$ can extremely approximate the real distribution $\mathcal{D}(X_s)$. Thus, $\mathcal{D}(\hat{X}_t)$  is relatively symmetric with $\mathcal{D}(X_s)$, making it easier to learn cross-domain correspondences between the source and target in the later stage. Oppositely, previous methods~\cite{pinkney2020resolution,richardson2021encoding, song2021agilegan} typically combine StyleGAN2 with inversion methods~\cite{abdal2020image2stylegan++,tov2021designing}, mapping source images to the $z$ space or $\mathcal{W}/\mathcal{W}+$ space of StyleGAN2 and leveraging this unconditional generator to synthesize the corresponding results. Therefore, it is hard to ensure that arbitrary portraits (i.e., out-of-domain images) can be embedded in the low-dimensional $z$ space or style-disentangled $\mathcal{W}/\mathcal{W}+$ space, due to the ``distortion-editability trade-off'' illustrated in \cite{tov2021designing, roich2021pivotal}. This inversion process leads to extra identity and structure details missing for image translation tasks.

\subsection{Geometry expansion module}
The previous module uses the source distribution as the ground-truth distribution to calibrate the target distribution. However, all images in the source domain (FFHQ) have been aligned with the standard facial position,  making the network heavily rely on the positional semantics for synthesis and further limit the network's capability to process real-world images. To release these constraints and support full-image inference stated in Section~\ref{sec:Inference}, we apply the geometry transformation $T_{Geo}$ to both source samples $\hat{x}_s/x_s$ and target samples $\hat{x}_t$, thus producing geometry extended samples $\tilde{x}_s$ and $\tilde{x}_t$. 
$T_{Geo}$ is performed with the random scale ratio $\mu\in[0.8, 1.2]$ and the random rotation angle $\gamma\in[-\frac{\pi}{2}, \frac{\pi}{2}]$.

% %%%%%%%%%%%% figure 5
\begin{figure}
\begin{center}
\setlength{\abovecaptionskip}{-0.1cm}
\includegraphics[width=0.95\linewidth]{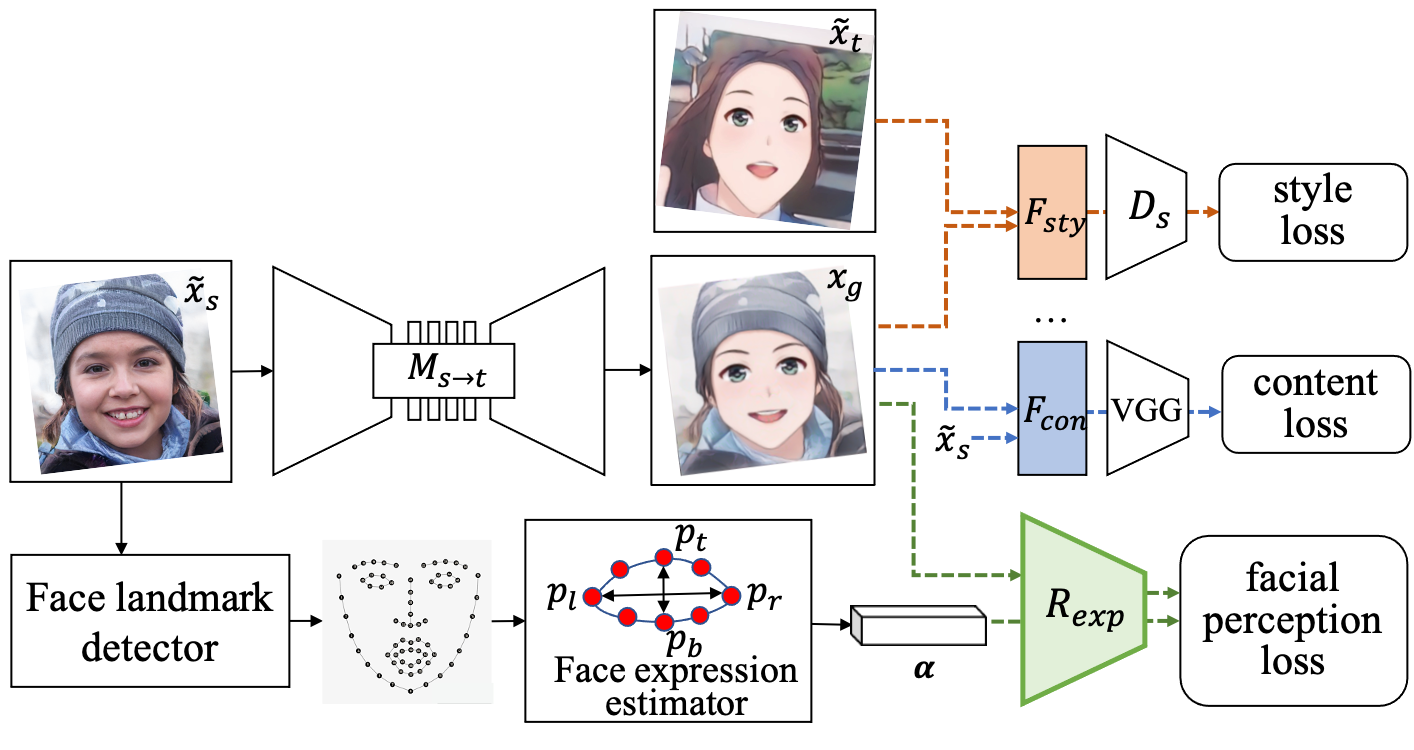}
\caption{The architecture of the texture translation network. }
\label{fig:ttn}
\end{center}
\end{figure}

\subsection{Texture translation network}
\label{sec:ttn}
The texture translation network (TTN) aims to learn cross-domain correspondences between the calibrated domains ($\tilde{X}_s, \tilde{X}_t$) in an unsupervised way. Although the first module can produce roughly aligned pairs by sampled noises $z$, it fails to preserve content details due to its nature of global mapping and also cannot handle arbitrary real faces with the additional inversion error. Considering sufficient texture information in reconstructed two domains but inaccurate texture mapping between them, we introduce a mapping network $\mathcal{M}_{s\to t}$ with the U-net architecture~\cite{ronneberger2015u} to convert the global domain mapping to the local texture transformation, thus learning a fine-grained texture translation in the pixel level. 

Due to the utilization of sufficient data of source photos, the reconstructed source distribution can extremely approximate the original source distribution ($D(X_s)\approx D(\hat{X_s})$). 
Thus, we directly use real sources (followed by geometry expansion) and calibrated target samples (after content and geometry calibration) for symmetric translation. In this process, the symmetric features are converted from the image level to the domain level. 
It is also worthy of noting that the proposed TTN is trained in an unsupervised way with unpaired images. Even when using real images as inputs, no inversion method is required to produce corresponding stylized samples. The style image $\tilde{x}_t \in \tilde{X}_t$ is randomly sampled and is only used to provide the style representation other than the ground truth, in order to get away with local optimum. 
It should be pointed out that we simply use the same sample for all modules in Figure 3 to make a concise and intuitive illustration of our method.

\subsubsection{Multi-representation constraints.}
%{\bf Multi-representation constraints.} 
 Inspired by the way of representation decomposition in~\cite{wang2020learning}, we extract the style representation $\mathcal{F}_{sty}$ from $\tilde{x}_t$ and $x_g$ via texture and surface decompositions, and use the discriminator ${D}_s$ to guide $\mathcal{M}_{s\to t}$ to synthesize $x_g$ in the similar style of $\tilde{x}_t$. The style loss $\mathcal{L}_{sty}$ is computed by penalizing the distance between the style representation distributions of real stylized images and generated images:
% equation 1
\begin{equation}
\begin{split}
\mathcal{L}_{sty}=
&\mathbb{E}_{\tilde{x}_s}[log(1-D_s(\mathcal{F}_{sty}(\mathcal{M}_{s\to t}(\tilde{x}_s) )))]\\
&+\mathbb{E}_{\tilde{x}_t}[log(D_s(\mathcal{F}_{sty}(\tilde{x}_t))].
\end{split}
\end{equation}
The pre-trained VGG16 network~\cite{simonyan2014very} is used to extract the content representations $\mathcal{F}_{con}$ from source images $\tilde{x}_s$ and generated images $x_g$ to ensure the content consistency. The content loss $\mathcal{L}_{con}$ is formulated as the L1 distance between $x_g$ and $\tilde{x}_s$ in the VGG feature space:
% equation 2
\begin{equation}
\mathcal{L}_{con} = || VGG(\tilde{x}_s), VGG(\mathcal{M}_{s\to t}(\tilde{x}_s)) ||_1.
\end{equation}

\subsubsection{Facial perception constraint.} To further encourage the network to produce stylized portraits with exaggerated structure deformations (such as the simplified mouth and big delicate eyes), an auxiliary expression regressor $\mathcal{R}_{exp}$ is introduced to guide the synthesis process. In other words, we inherently impulse local structure deformations by constraining the facial expression of synthetic images via $\mathcal{R}_{exp}$, which pays more attention to the region of facial components (e.g., mouth and eyes). Specifically, $\mathcal{R}_{exp}$ consists of $n$ regression heads on top of the feature extractor $\mathcal{E}_f$, where $n$ denotes the number of expression parameters. Both $\mathcal{E}_f$ and ${D}_s$ follow the PatchGAN architecture~\cite{isola2017image}.
To achieve a faster training procedure, we directly apply the learned regressor to estimate the expression scores of generated images $x_g$. The facial perception loss is calculated by: 
% equation 3
\begin{equation}
\mathcal{L}_{per} = || \mathcal{R}_{exp}(x_g) - \bm{\alpha} ||_2,
\end{equation}
where $\bm{\alpha}={\bm{\alpha}_1,..,\bm{\alpha}_n}$ denotes the expression parameters extracted from the source image $\tilde{x}_s$. We set $n=3$ and define $\bm{\alpha}_i$ $\in [0, 1]$ as the opening degrees of the left eye, the right eye, and the mouth, respectively. With the facial points $p$ extracted from $\tilde{x}_s$, 
 $\bm{\alpha}_i$ can be easily obtained by calculating the height-to-width ratio of the bounding box of specific facial components.

% %%%%%%%%%%%% figure 6
\begin{figure}
\begin{center}
\setlength{\abovecaptionskip}{0cm}
\includegraphics[width=0.85\linewidth]{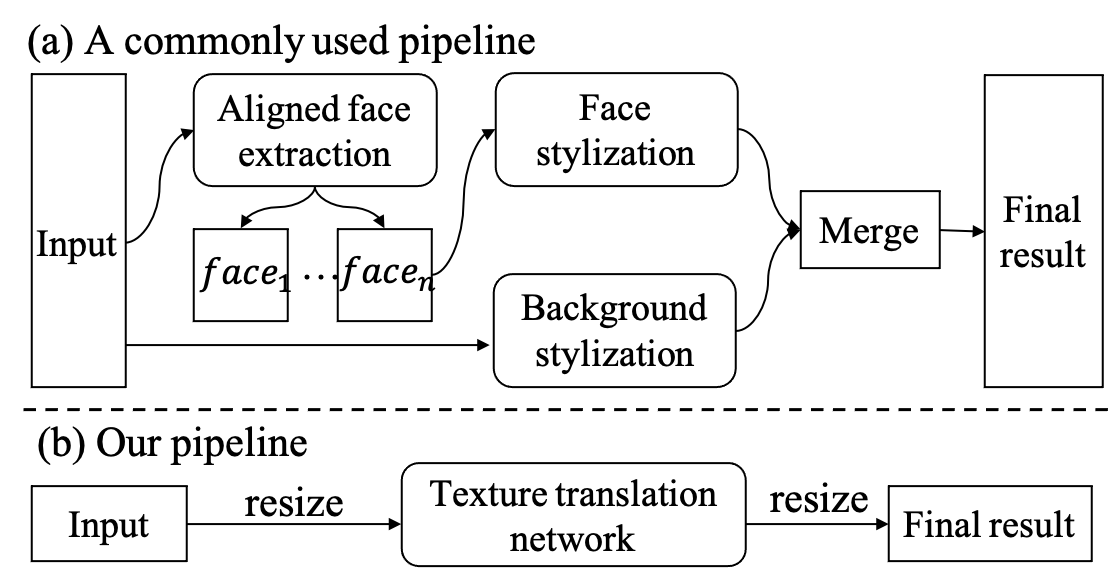}
\caption{Pipelines of full image translation. Instead of exploiting complicated architectures as other existing approaches, we achieve the goal in an elegant single network with one evaluation.}
\label{fig:inference}
\end{center}
\end{figure}

\subsubsection{Training.} Given $\tilde{x}_s$ and $\tilde{x}_t$ from the calibrated source and target domains, the texture translation model is trained with the full loss function consisting of a style term, a content term, a facial perception term, and a total-variation term:
% equation 4
\begin{equation}
\mathcal{L}_{total} = \mathcal{L}_{sty}+\lambda_{con}\mathcal{L}_{con}+\lambda_{per}\mathcal{L}_{per}++\lambda_{tv}\mathcal{L}_{tv},
\end{equation}
where $\lambda$ denotes the weight of each corresponding loss. The total-variation loss $\mathcal{L}_{tv}$ is used to smooth the generated image $x_g$, which can be computed by:
% equation 5
\begin{equation}
\mathcal{L}_{tv} = \frac{1}{h*w*c}||\nabla_u(x_g)+\nabla_v(x_g)||,
\end{equation}
where $u$ and $v$ denote horizontal and vertical directions, respectively.

\subsection{Inference}
\label{sec:Inference}

Different from previous works~\cite{Kim2020U-GAT-IT, song2021agilegan} that are limited to aligned face stylization, our model enables full-image rendering for arbitrary portrait images containing multiple faces in rotations. A common practice to achieve the aforementioned goal is to process face and background independently. As described in Figure~\ref{fig:inference}, they firstly extract aligned faces from the input image and stylize all the faces one-by-one. Then, the background image is rendered with some specialized algorithms and merged with stylized faces to obtain the final result. Instead of using such complex pipeline, we found that our texture translation network can directly render stylized results from full images in one-pass evaluation. With domain-calibrated images, the network sees the entire texture contents during training, so it implicitly encodes the contextual information of the background as well as the facial appearance. Combined with the geometry expansion module, it is scale and rotation invariant against raw face processing. Since a range of the scale ratio is adopted in GEM, input images are all resized to scales that can be satisfactorily handled. We experimentally found that images with the resolution lower than $2K\times2K$ can be well handled with no blur appearing in our synthesized images.

\section{Experimental Results}

% %%%%%%%%%%%% figure 7
\begin{figure}
\begin{center}
\setlength{\abovecaptionskip}{-0.1cm}
\includegraphics[width=1.0\linewidth]{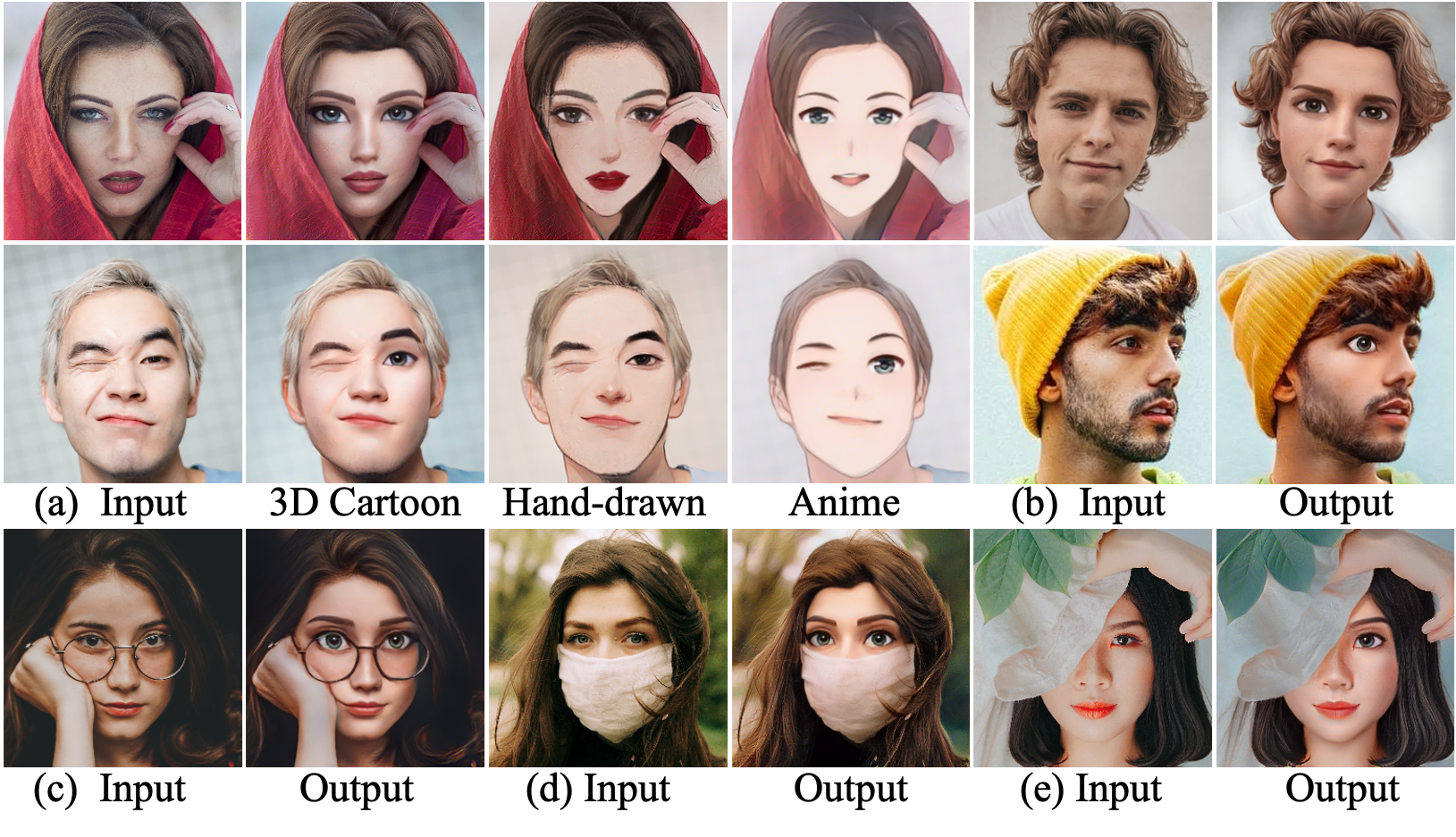}
\caption{Results of synthesized portraits in various styles (a) and complicated scenes (b-e), due to occlusions, accessories, or poses.}
\label{fig:result}
\end{center}
\end{figure}

 \subsection{Implementation Details}
Regarding the training process of our overall network, we first train CCN with the loss function described in the supplemental material, and use its inference phase to calibrate contents. Then, GEM is directly adopted for geometric calibration without training. Finally, with calibrated domains, TTN is trained with the loss function introduced in Section~\ref{sec:ttn}. Specifically, for CCN, the weights of $G_t$ and $D_t$ are initially retrieved from the StyleGAN2 config-f $256\times256$ FFHQ model and fine-tuned following~\cite{karras2020analyzing}. We set $k=4$ to blend the model. Content calibrated samples $\hat{x}_t$ are shuffled with raw samples $x_t$ and they are processed by $T_{Geo}$ to obtain final calibrated samples $\tilde{x}_t$. Specifically, the number of generated style samples is 10,000. 
For the training process of TTN, we use 10,000 images from FFHQ processed by $T_{Geo}$ as calibrated source photos $\tilde{x}_s$ and mixed data consisting of real images ($\sim$100) and generated samples (10,000) as target exemplars $\tilde{x}_t$. 
$\mathcal{F}_{con}$ is extracted from the layer $l=conv\{4\_4\}$ of the pre-trained VGG16 model. $\mathcal{R}_{exp}$ is trained with labeled attributes, which are computed by combining existing face landmark detectors~\cite{zhang2016joint, animeface}. With the learned $\mathcal{R}_{exp}$, we adopt the Adam optimizer~\cite{kingma2014adam} with $\beta_1 = 0.5$ and $\beta_2 = 0.99$ to train the TTN model for around 10k iterations. The learning rate is set to $1\times10^{-4}$ and $(\lambda_{con}, \lambda_{per}, \lambda_{tv})$ is set to $(2\times10^2, 1, 10^4)$. The training flow and hyper-parameters involved are the same for all styles.

\subsection{Datasets}
For source photos, we use 10,000 images from the FFHQ dataset~\cite{karras2019style} as the training data. For target exemplars, we collect several art portrait assets (e.g., 3d cartoon, hand-drawn, barbie, comic, etc.) from the Internet and each asset contains approximate 100 images for a similar style. Only the anime style asset is created by artists and other assets are randomly downloaded from the websites. For the evaluation, we use the first 5,000 images of the CelebA dataset~\cite{CelebAMask-HQ} for testing. 
\subsection{Artistic portrait generation}
 \emph{The ability to synthesize high-preserving contents.} 
Besides test cases in CelebA, we also validate the capability of our model by stylizing wild portrait images collected from the Internet. As shown in Figure~\ref{fig:result}, not only the global structures between the input and the output are consistent, but also the local details such as accessories, background, and identity are highly preserved. 

 \emph{The generality to handle complicated scenes.} To verify the strong generality of our model to handle complex real-world scenes, we test our model with hard cases, which contain heavy occlusions (Figure~\ref{fig:result} (c, d, e)) and rare poses (b). Our method shows high robustness for these cases. 
We also provide more results of our method with diverse inputs (e.g., different skin tones) in Figure~\ref{fig:res_diverse}.

 \emph{The scalability to transfer various styles.} With limited exemplars of a new style, this unified framework can be directly used to train a new style model. We show stylized results (e.g., 3D cartoon, hand-drawn, anime) produced by different style models in Figure~\ref{fig:result} (a) and more results are provided in supplemental materials (Supp).

% %%%%%%%%%%%% figure 8
\begin{figure}
\begin{center}
\setlength{\abovecaptionskip}{-0.1cm}
\includegraphics[width=0.98\linewidth]{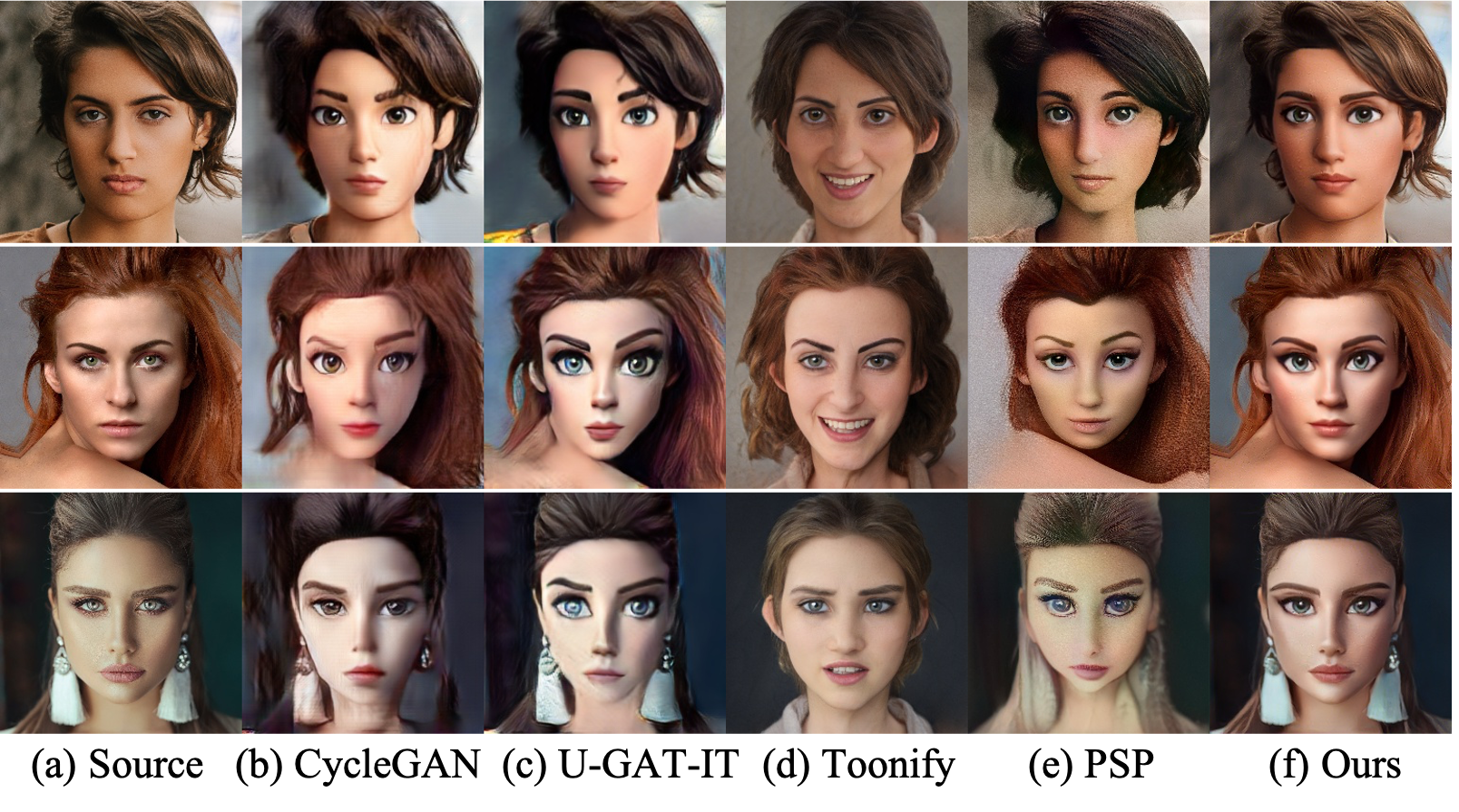}
\caption{Qualitative comparison with four state-of-the-art methods. }
\label{fig:compare}
\end{center}
\end{figure}

% %%%%%%%%%%%% figure 9
\begin{figure}
\begin{center}
\setlength{\abovecaptionskip}{-0.1cm}
\includegraphics[width=0.98\linewidth]{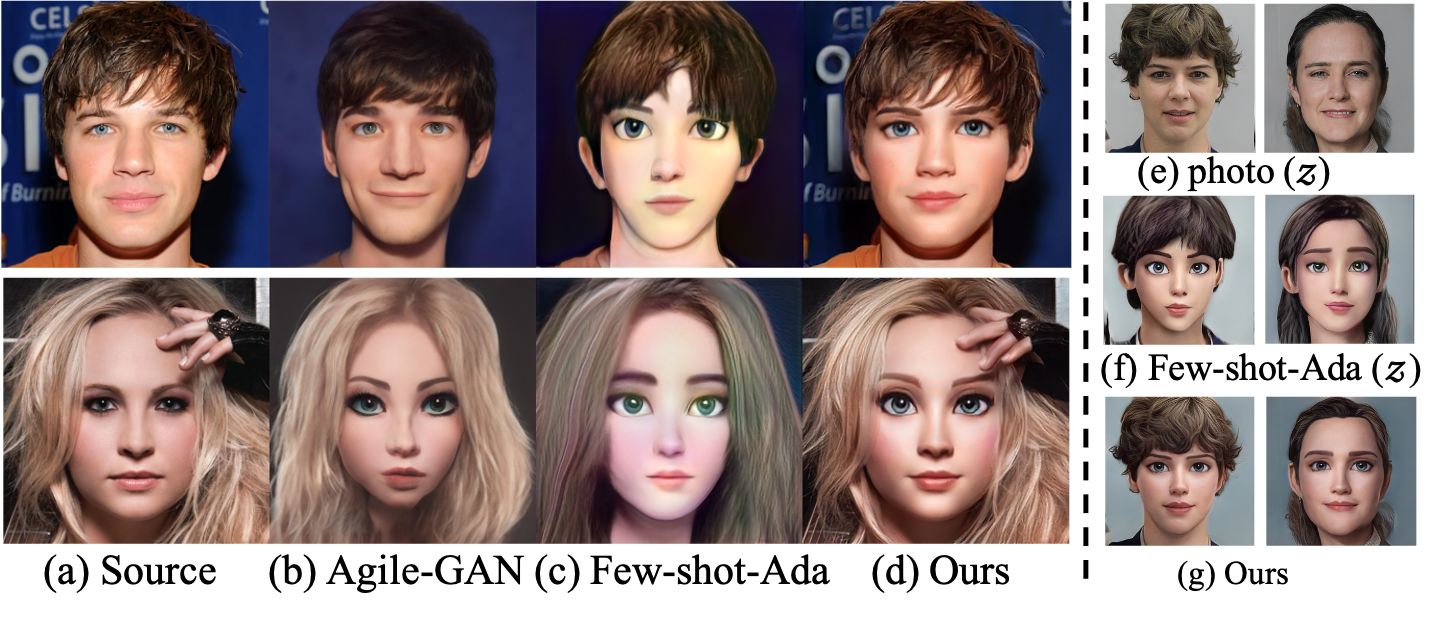}
\caption{Qualitative comparison with AgileGAN and Few-shot Ada.}
\label{fig:compare2}
\end{center}
\end{figure}

\subsection{Comparison with the state of the art}
\subsubsection{Qualitative comparison.} In Figure~\ref{fig:compare} and Figure~\ref{fig:compare2}, we compare the synthetic results of our method with six state-of-the-art head cartoonization methods which can be categorized into two types; a) image-to-image translation based methods: CycleGAN~\cite{zhu2017unpaired}, U-GAT-IT~\cite{Kim2020U-GAT-IT}; b) StyleGAN-adaption based methods: Toonify~\cite{pinkney2020resolution}, PSP~\cite{richardson2021encoding}, AgileGAN~\cite{song2021agilegan} and Few-shot-Ada~\cite{ojha2021few}. For the first four methods, the results are produced by directly using source codes or trained models released by authors. For AgileGAN, since its code and trained model are not publicly available, we directly evaluate our method using examples provided by them officially. 
Few-shot-Ada is an unconditional generative model and can only synthesize <photo, cartoon> pairs, with the same random noise fed into its source and adapted generators respectively. 
So we use the inversion algorithm in ~\cite{karras2020analyzing} to project real faces to the latent space and use their adapted generator to produce stylized results for arbitrary images (Figure~\ref{fig:compare2} (c)). 
Considering the inversion error, we also test their method in the noise manner and use their synthesized images as arbitrary inputs for our method (Figure~\ref{fig:compare2} (e, f, g)). As we can see, our method still outperforms it with more content details. Compared with other approaches, our method produces more realistic results in both content similarity and style faithfulness. The facial identity is better preserved and even detailed accessories or extra body parts are successfully synthesized. More comparison results can be found in Supp.

\subsubsection{Quantitative comparison.}
We evaluate the quality of our results using the Frechet Inception Distance (FID) metric~\cite{heusel2017gans}, which is a common metric to measure the visual similarity and distribution discrepancy between two sets of images. We generate stylized images from the CelebA dataset for each method, and compute their FID value from the training cartoon dataset. 
To further evaluate the identity similarity (ID) between generated and source images, we extract identity vectors using a pre-trained face recognition model~\cite{wang2018cosface} and adopt the normalized cosine distance to measure the similarity. 
As shown in Table~\ref{tab:quantitative}, our method generates not only more realistic details with the lowest FID value, but also more similar identity with the highest ID value.

\begin{table}
\setlength{\abovecaptionskip}{0cm}
\setlength{\belowcaptionskip}{-0.1cm}
  \centering
    \caption{Quantitative comparison of our method and four state-of-the-art approaches evaluated by two metrics (i.e., FID and ID) and user studies.}
    \small
     \resizebox{0.98\linewidth}{!}{
  \begin{tabular}{ccccc}
    \toprule
    Method &  FID $\downarrow$  &  ID $\uparrow$  & Pref. A $\uparrow$ & Pref. B$\uparrow$  \\
    \midrule
    CycleGAN~\cite{zhu2017unpaired} & 57.08 & 0.55 & 7.1  & 1.4 \\
     Ugatit~\cite{Kim2020U-GAT-IT} & 68.40 & 0.58  & 5.0  & 1.5 \\
    Toonify~\cite{pinkney2020resolution} & 55.27  & 0.62 & 3.7  & 4.2 \\
    PSP~\cite{richardson2021encoding} & 69.38  & 0.60 & 1.6  & 2.5  \\
    Ours & \textbf{35.92}  & \textbf{0.71} & \textbf{82.6} & \textbf{90.5}  \\
    Ours-w/o CCN & {58.52}  & {0.58} & - & -  \\
    Ours-w/o GEM & {37.46}  & {0.70} & - & -  \\
    Ours-w/o TTN & {39.68}  & {0.59} & - & -  \\
    \bottomrule
  \end{tabular}}
  \label{tab:quantitative}
\end{table}

\subsubsection{User study.}
As portrait stylization is often regarded as a subjective task, we resort to user studies to better evaluate the performance of the proposed method. We conducted two user studies on the results in terms of the stylization effects and faithfulness to content characteristics. In the first study, participants were asked to select the best stylized images with less distorted artifacts (Pref. A). In the second study, participants were asked to point out which stylized images best preserve the corresponding contents (Pref. B). Each participant was shown 25 questions randomly selected from a question pool containing 100 examples for each study. In each question, we show an input source following by four stylized results of competing methods and ours, where the images are arranged in a random order. We receive 1,000 answers from 40 subjects in total for each study. 
 As shown in Table~\ref{tab:quantitative}, over 80\% of our results are selected as the best in both two metrics, which proves a significant quality boost in stylization effects and faithful transfer obtained by our approach.

% %%%%%%%%%%%% figure 10
\begin{figure}
\begin{center}
\setlength{\abovecaptionskip}{-0.1cm}
\setlength{\belowcaptionskip}{-0cm}
\includegraphics[width=1\linewidth]{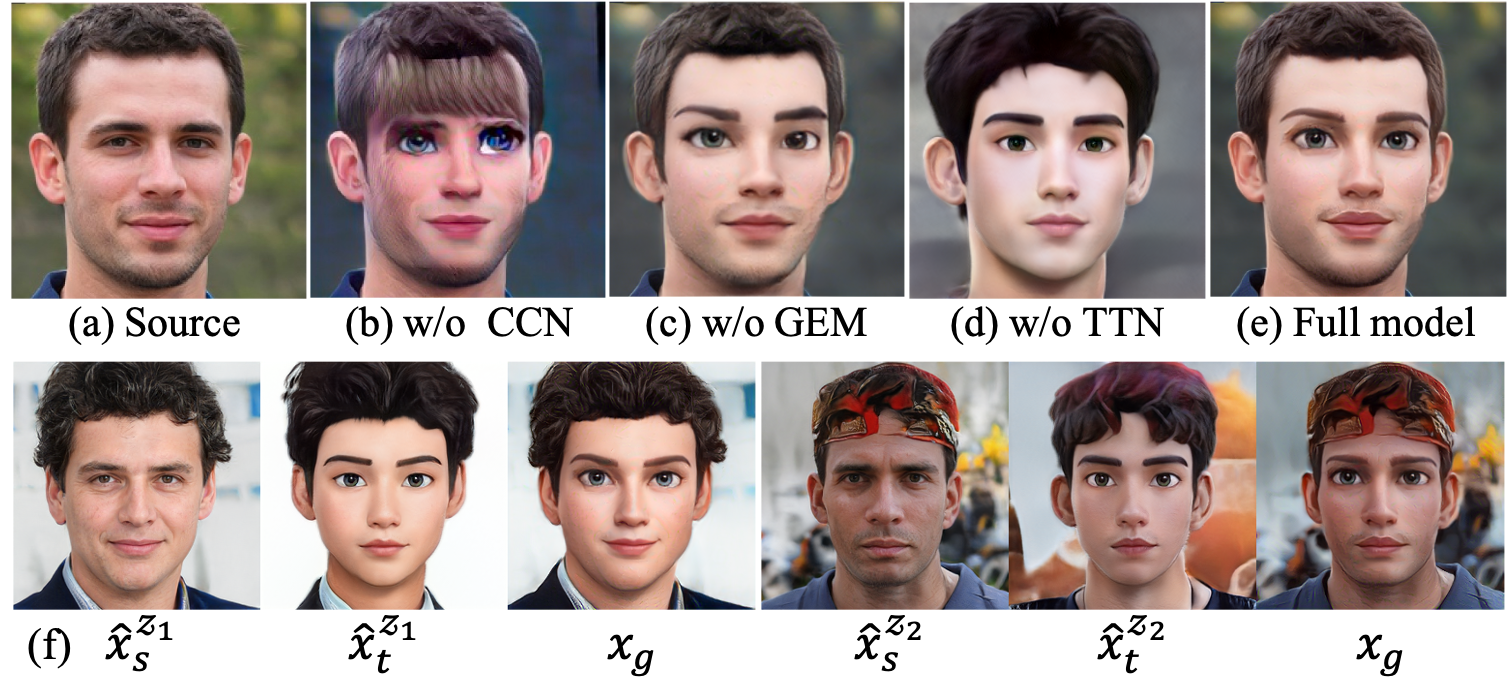}
\caption{Effects of our proposed CCN, GEM, and TTN.}
\label{fig:ablation_p1}
\end{center}
\end{figure}

 % %%%%%%%%%%%% figure 12
\begin{figure}
\begin{center}
\setlength{\abovecaptionskip}{-0.1cm}
\setlength{\belowcaptionskip}{0.1cm}
\includegraphics[width=0.87\linewidth]{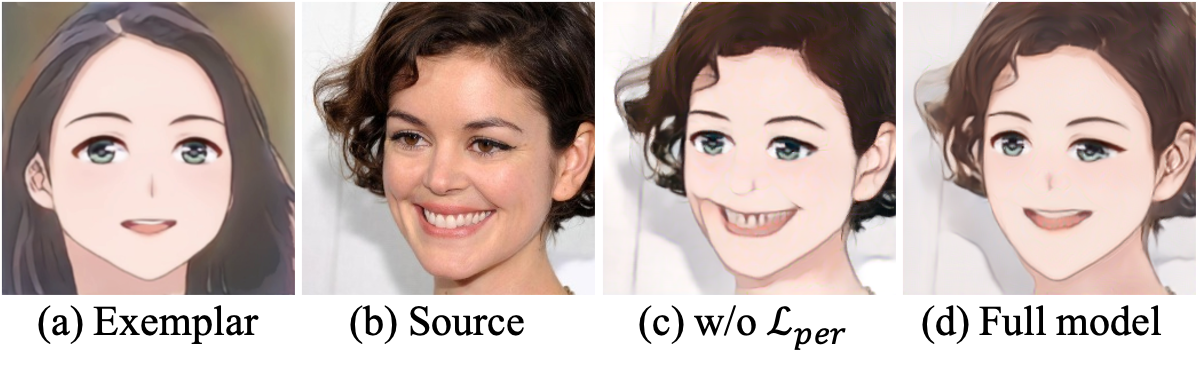}
\caption{Auxiliary effects of the facial perception loss. }
\label{fig:ablation_p2}
\end{center}
\end{figure}

\subsection{Ablation study}
\label{sec:ablation}

\subsubsection{The proposed three modules. } To verify the effectiveness of the proposed CCN, GEM, and TTN, we evaluate the performance of several variants of our method by removing each module independently. The qualitative and quantitative results are shown in Figure~\ref{fig:ablation_p1} and Table~\ref{tab:quantitative}, respectively. 
Our method w/o CCN can easily suffer from texture artifacts because of overfitting. CCN brings better generalization ability for this transfer task since it improves the diversity of target samples and calibrates the target distribution closer to the original source distribution. 
GEM makes our model more stable to the face alignment error and impulses full translations in freely spatial conditions. It is also necessary for the application of full-body image stylization in Section~\ref{sec:app_full}. 
For our method w/o TTN, we use the inversion method in~\cite{karras2020analyzing} to project real faces to the latent code and use $G_t$ of CCN to produce stylized results (Figure~\ref{fig:ablation_p1} (d)).  
Results of our method w/o TTN suffer from the content missing problem especially for arbitrary real faces out-of-domain. 
This stems from not only the GAN inversion error but also the function change in the domain adaption process. 
 To prove this, we show samples $(\hat{x}_s^{z}, \hat{x}_t^{z})$ produced by CCN with random noise $z$ in Figure~\ref{fig:ablation_p1} (f), and we can see that the issue is alleviated but still exists without the inversion process. 
 This is also an inherent problem along with all StyleGAN-adaption based methods ~\cite{song2021agilegan, ojha2021few, richardson2021encoding}. We tackle this problem with TTN and the results are shown in Figure~\ref{fig:ablation_p1} (e, f).  TTN significantly improves the network's ability of content preservation and makes it be capable of stylizing arbitrary real photos with more similar identities as the original ones.

\subsubsection{Facial perception loss.}
Due to the translation network’s strong ability of content preservation, it is difficult to achieve extremely exaggerated deformations, such as simplified noses and mouths in the anime style (see Figure~\ref{fig:ablation_p2} (c)). Actually, there is a trade-off between content similarity and style faithfulness. Here, we introduce the facial perception loss $\mathcal{L}_{per}$ to encourage large structure changes for local components (e.g., eyes, nose, and mouth) and unchanged structures for other components, thus achieving adaptive deformation for different parts (see Figure~\ref{fig:ablation_p2} (d)). It is worthy of noting that $\mathcal{L}_{per}$ is designed exclusively for extremely exaggerated styles, the proposed method can still produce satisfactory results for undeformable styles without $\mathcal{L}_{per}$.

% %%%%%%%%%%%% figure 11
\begin{figure*}
\begin{center}
\setlength{\abovecaptionskip}{-0.1cm}
\setlength{\belowcaptionskip}{0.1cm}
\includegraphics[width=1\linewidth]{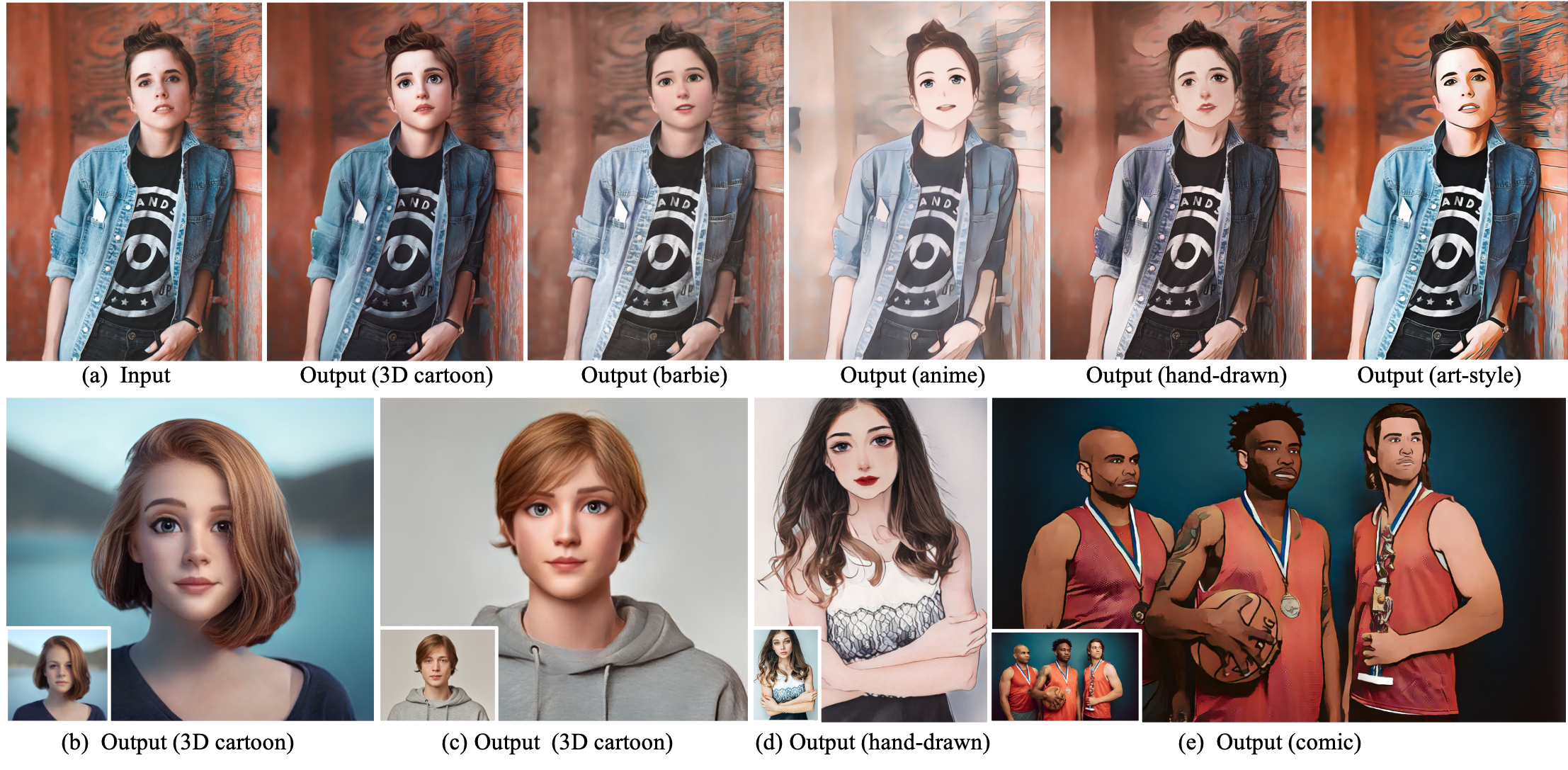}
\caption{Results of stylized full images in various styles (a) and casual style cases (b, c, d, e) with source images in the bottom left corner.}
\label{fig:res_full}
\end{center}
\end{figure*}

\subsection{Full-body image translation}
\label{sec:app_full}
Given training samples observed only in the head region, we find that our model achieves can also achieve full-body image translation in one evaluation with a single network. We show full-body results with various styles and some random cases in Figure~\ref{fig:res_full}. As we can see, the proposed method works well for arbitrary images with harmonious tones and adaptive deformations (e.g., the exaggerated eyes and faithful body). More synthesis results and some failure cases of our method can be found in Supp.

 % %%%%%%%%%%%% figure 13
\begin{figure*}
\begin{center}
\setlength{\abovecaptionskip}{-0.1cm}
\setlength{\belowcaptionskip}{0.1cm}
\includegraphics[width=0.84\linewidth]{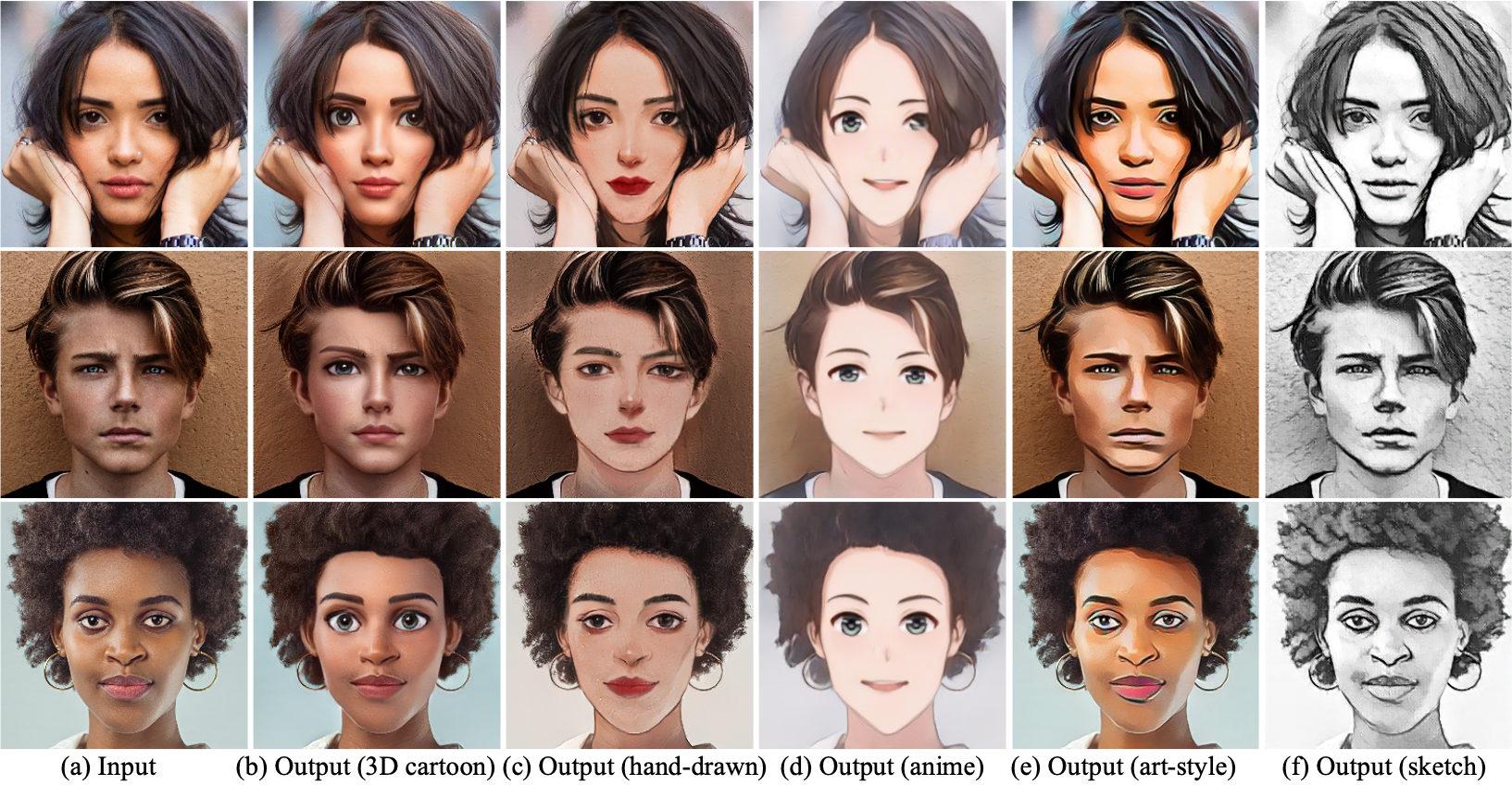}
\caption{Results of stylized portraits with diverse input images.}
\label{fig:res_diverse}
\end{center}
\end{figure*}

%\begin{figure}
%\begin{center}
%\setlength{\abovecaptionskip}{-0.1cm}
%\setlength{\belowcaptionskip}{0cm}
%\includegraphics[width=0.87\linewidth]{pic//limitation.png}
%\caption{Failure cases due to the disturbed illumination.}
%\label{fig:limit}
%\end{center}
%\end{figure}

%\section{Limitations}
%Because of the inherent characteristics of some styles (i.e., hand-drawn and anime), our synthesized results might not be natural enough when there exists server lighting shadows in human faces, as shown in Figure~\ref{fig:limit}. But some styles (i.e., 3D cartoon) can still be well handled owing to its specific nature. 

\section{Conclusion}
We presented DCT-Net, a novel framework for stylized portrait generation, which not only makes a boost in ability, generality, and scalability for the head stylization task, but also achieves effective full-body image translation in an elegant manner. Our key idea is to calibrate the biased target domain firstly, and learn a fine-grained translation later. Specifically, the content calibration network was introduced for diverse textures and the geometry expansion module was designed to release spatial constraints. With calibrated samples produced by the above two modules, our texture translation network easily learns cross-domain correspondences with delicately designed losses. Experimental results demonstrated the superiority and effectiveness of our method. We also believed that our solution of domain-calibrated translation could inspire future investigations on image-to-image translation tasks with biased target distribution. 

% Bibliography
\bibliographystyle{ACM-Reference-Format}
\bibliography{sample-bibliography}

\end{document}